\documentclass[letterpaper, 10 pt, conference]{ieeeconf}

\IEEEoverridecommandlockouts
\usepackage{amsmath,amssymb}
\usepackage{graphicx}
\usepackage{placeins}
\usepackage{capt-of}
\usepackage{booktabs}
\usepackage{adjustbox}

\title{\LARGE \bf
Dual-Stream Memory for Vision-Language-Action Policies
}

\author{\authorblockN{Yang Wang$^{1}$, Ping Liu$^{1}$, Ruiyu Wang$^{1,2}$, Xuhua Chen$^{1}$, Lingfeng Zhang$^{1}$, and Tao Zhang\textsuperscript{1}\authorrefmark{2}}
\authorblockA{$^{1}$Magic-Lab Team, Magiclab Robotics Inc.\\
$^{2}$University of California, Los Angeles, USA}}

\makeatletter
\newcommand{\papertablecaption}[2]{%
  \begingroup
  \long\def\@makecaption##1##2{%
    {\small\noindent TABLE~\thetable: ##2\par}\vskip6pt}%
  \captionof{table}{#1}\label{#2}%
  \endgroup}
\makeatother

\begin{document}
\raggedbottom

\twocolumn[{
\begin{@twocolumnfalse}
\maketitle
\begin{minipage}{\textwidth}
\centering
\includegraphics[width=0.94\textwidth]{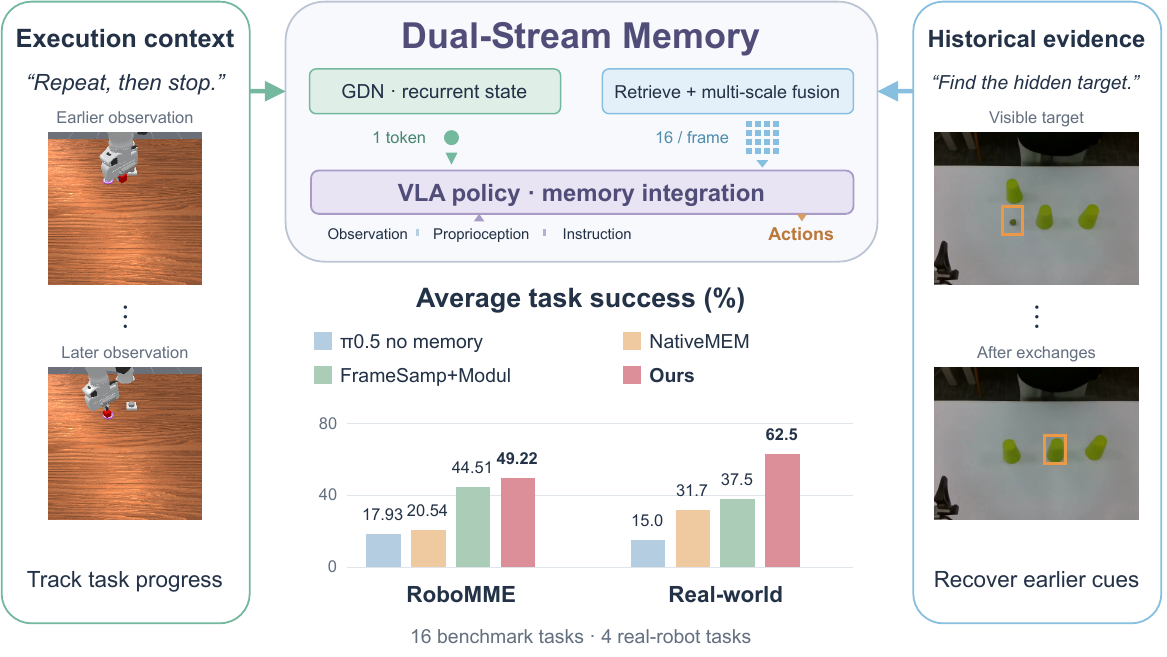}
\captionof{figure}{Dual-stream memory combines recurrent execution context with retrieved spatial evidence. Bars show mean task success rate (TSR) across sixteen RoboMME tasks and four real-world tasks evaluated on the Piper robot (Sections~\ref{sec:full} and~\ref{sec:real}). Images illustrate a RoboMME episode and a human ShellGame demonstration. Orange boxes mark the visible target and its final covering cup; memory symbols are schematic.}
\label{fig:teaser}
\end{minipage}
\vspace{10pt}
\end{@twocolumnfalse}
}]
\begingroup
\renewcommand{\thefootnote}{}%
\footnotetext{\authorrefmark{2}Corresponding author.}%
\endgroup
\thispagestyle{empty}
\pagestyle{empty}

\begin{abstract}
Memory-dependent manipulation requires a policy to track execution state and recover visual evidence that is no longer observable. We propose dual-stream memory for vision-language-action policies, assigning these requirements to complementary representations. A task-agnostic Gated DeltaNet compresses history into one temporal token, supervised by progress prediction and action-conditioned feature reconstruction. A task-conditioned retriever retains frames above a fixed relevance threshold; linear multi-scale fusion encodes each frame into sixteen spatial tokens. We then freeze the temporal compressor and jointly optimize retrieval and policy learning. Across sixteen RoboMME tasks, the complete model achieves 49.22\% mean success versus 44.51\% for the published FrameSamp+Modul reference, improving fourteen tasks and all four suite means. On four real-robot tasks, mean success reaches 62.5\% versus 37.5\%, with the largest gains on repeated actions requiring execution-stage discrimination. These results support combining execution context with selective visual access, while failures under cup swaps expose persistent identity tracking as an unresolved requirement.
\end{abstract}

\noindent\textit{Index Terms}---Vision-language-action models, robot memory, recurrent compression, task-conditioned retrieval.

\section{INTRODUCTION}

\begin{figure*}[t]
\centering
\includegraphics[width=\textwidth]{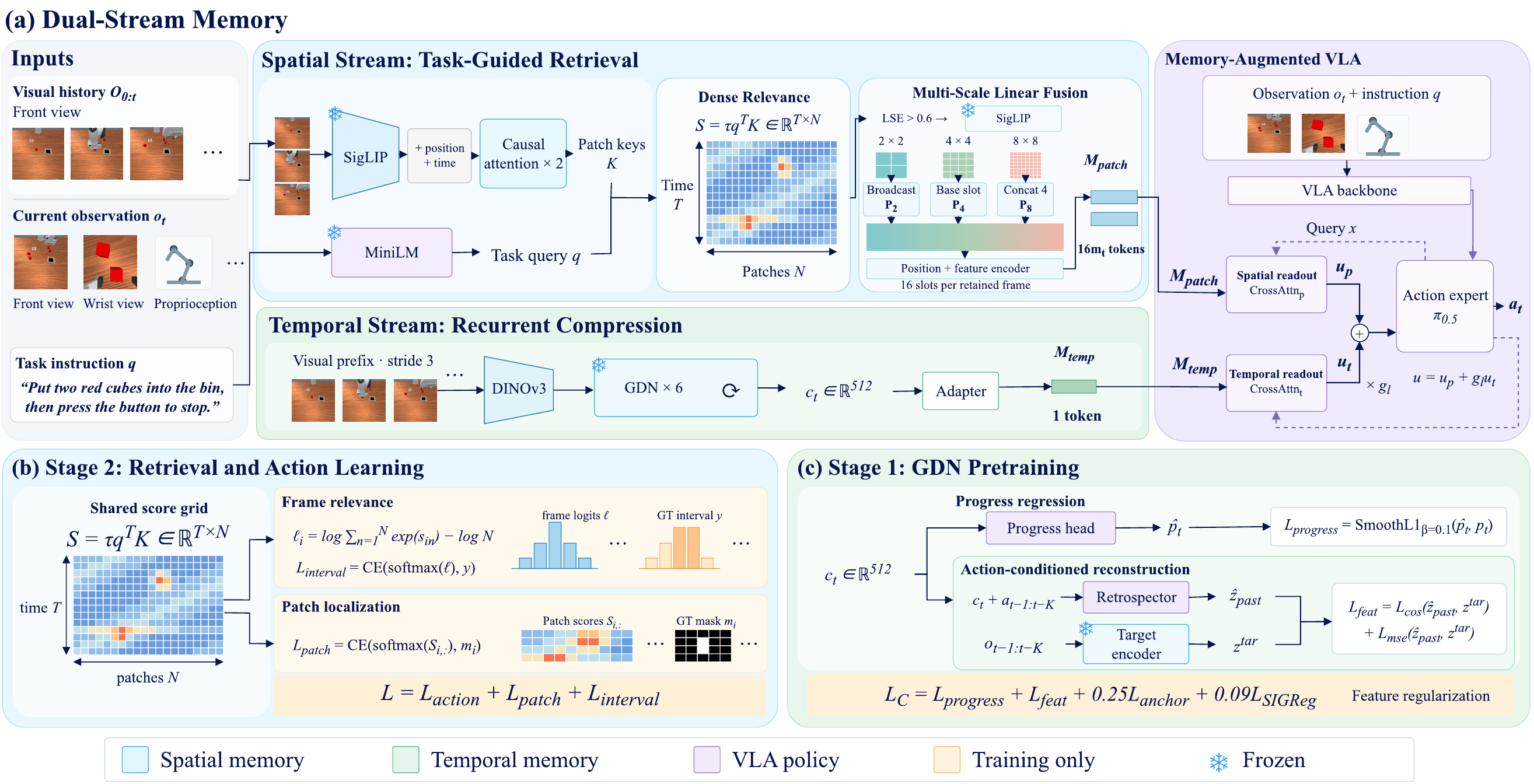}
\caption{Dual-stream memory. (a) GDN compresses visual history into one temporal token; threshold retrieval and multi-scale fusion produce $16m_t$ spatial tokens from $m_t$ retained frames. Dashed arrows denote action-expert queries to the memory readouts. Their gated sum modulates the action expert. (b) Stage 2 combines action and retrieval losses, with GDN frozen. (c) Stage 1 pretrains GDN with compression objectives; its auxiliary heads are then removed. Snowflakes mark frozen modules in their respective stages. Camera images show a RoboMME episode; score displays are schematic.}
\label{fig:pipeline}
\end{figure*}

Long-horizon manipulation can produce perceptually similar observations that require different actions, depending on prior execution or earlier visual evidence. Repeated actions require execution-stage information, whereas occlusion and target reidentification require distinguishing visual detail that may no longer be observable. These requirements motivate separate representations for evolving execution context and accessible historical evidence (Fig.~\ref{fig:teaser}).

Vision-language-action (VLA) models connect visual observations and instructions to robot actions~\cite{brohan2023rt2,kim2024openvla,ghosh2024octo,black2025pi0,black2025pi05}. Large-scale robot data, action chunking, and generative action models broaden their capabilities~\cite{brohan2022rt1,oxe2023,chi2023diffusion,zhao2023act}. However, they do not determine which episodic information a policy should retain. Recent memory-augmented policies preserve perceptual observations, cognitive summaries, or representations at several temporal scales~\cite{shi2026memoryvla,torne2026mem,wang2026nativemem}. Which historical information should be compressed, and which should remain directly accessible?

Recurrent compression summarizes execution history, but a compact state need not preserve detail for later target identification. Storing historical features preserves access, but uniform sampling may miss brief evidence, and spatial pooling may discard distinguishing detail. We therefore distinguish execution context, frame relevance, and within-frame detail.

Our design combines recurrent temporal memory with selected spatial evidence (Fig.~\ref{fig:pipeline}). We make three contributions:
\begin{enumerate}
\setlength{\itemsep}{0pt}
\renewcommand{\labelenumi}{\arabic{enumi})}
\item We introduce a task-agnostic Gated DeltaNet (GDN) compressor supervised by progress prediction and action-conditioned feature reconstruction. These objectives encourage execution context and reconstructable visual history in one temporal token.
\item We formulate task-conditioned evidence access as threshold retrieval with shared temporal and spatial supervision. The retained-frame count adapts to the available evidence.
\item We introduce linear multi-scale fusion to combine coarse context and fine spatial detail within sixteen tokens per retained frame.
\end{enumerate}

\section{RELATED WORK}

\subsection{Memory for robot policies}
Robot memory research distinguishes sequencing skills from retaining information within an episode. CALVIN and LIBERO study sequential manipulation and knowledge transfer~\cite{mees2022calvin,liu2023libero}; RoboMME targets Counting, Permanence, Reference, and Imitation~\cite{dai2026robomme}. Its FrameSamp+Modul baseline exposes sampled historical features to a VLA policy.

Beyond sampled observations, memory interfaces compress and organize history. Latent bottlenecks and cross-attention provide mechanisms for reading visual context~\cite{jaegle2021perceiver,alayrac2022flamingo}. MemoryVLA combines perceptual and cognitive memory, MEM organizes video and text at multiple temporal scales, and NativeMEM learns policy-aligned memory tokens~\cite{shi2026memoryvla,torne2026mem,wang2026nativemem}. Our design separates recurrent state from retrieved evidence, with distinct operations for execution context, frame selection, and spatial detail.

\subsection{Recurrent compression and selective access}
Content-based attention reads stored representations~\cite{vaswani2017attention}; Transformer-XL, Compressive Transformer, and Recurrent Memory Transformer extend historical access through recurrence or compression~\cite{dai2019transformerxl,rae2019compressive,bulatov2022rmt}. Linear attention and state-space models instead maintain recurrent states~\cite{katharopoulos2020linear,gu2023mamba,dao2024mamba2}. DeltaNet introduces delta-rule updates, while GDN adds data-dependent forgetting~\cite{yang2024deltanet,yang2025gateddelta}. Test-Time Training and RoboTTT further explore expressive recurrent states and robot-policy context~\cite{sun2024ttt,jiang2026robottt}. We build on GDN by supervising execution progress and reconstructable visual history, with a separate spatial memory bank for direct evidence access.

\subsection{Spatial representations for memory}
Patch encoders provide spatial features~\cite{dosovitskiy2020vit}: DINO learns self-supervised visual structure~\cite{caron2021dino,oquab2023dinov2,simeoni2025dinov3}, whereas CLIP and SigLIP learn image--text correspondence~\cite{radford2021clip,zhai2023siglip}. Feature pyramids, hierarchical Transformers, and mixed-resolution tokenization preserve information across spatial scales~\cite{lin2016fpn,liu2021swin,ronen2023mixedres}; PerAct emphasizes spatial structure for manipulation~\cite{shridhar2022peract}. We use DINOv3 for temporal compression and frozen SigLIP for retrieval and spatial memory. Our fusion aligns coarse, base, and fine features within sixteen slots, enriching retained frames without enlarging their token budget.

\section{METHOD}
\subsection{Dual-stream memory}
Let $O_{0:t}$ denote the front-view visual prefix available at decision step $t$. The temporal stream compresses a stride-sampled prefix into $c_t$, while the spatial stream selects and encodes task-relevant frames. Both streams are causal. The policy also receives current camera observations, proprioception, and the instruction. DINOv3 supplies one global token per frame for recurrent compression; frozen SigLIP supplies language-aligned patch features for evidence localization and spatial memory.

\subsection{Recurrent execution context}
\label{sec:gdn}
Progress supervision encourages the recurrent state to distinguish execution stages when repeated actions produce similar observations. Because a scalar progress target cannot capture earlier appearance, action-conditioned feature reconstruction supplies a complementary constraint on historical visual information.

The temporal stream processes front-view history at stride three. DINOv3 class tokens are projected to width 512 and passed through a six-layer GDN~\cite{yang2025gateddelta}. The final output yields $c_t\in\mathbb{R}^{512}$. A progress head predicts normalized execution progress $p_t=(t-d_e)/\max(T_e-d_e-1,1)$, where $d_e$ is execution onset and $T_e$ is episode length. Only execution frames supervise this head, using SmoothL1 loss with $\beta=0.1$. Episode endpoints define training targets and are not required at inference.

An auxiliary Retrospector combines $c_t$ with reverse-ordered executed actions to reconstruct past features from a frozen target encoder. It uses a four-layer causal Transformer and up to sixteen actions, sampled at gaps of 1--5 frames. Cosine and mean-squared losses on normalized features define $\mathcal{L}_{\mathrm{feat}}$. The joint compression objective is
\begin{equation}
\mathcal{L}_{\mathrm C}=\mathcal{L}_{\mathrm{progress}}+\mathcal{L}_{\mathrm{feat}}
+0.25\mathcal{L}_{\mathrm{anchor}}+0.09\mathcal{L}_{\mathrm{SIGReg}}.
\end{equation}
The anchor loss keeps online features close to frozen targets. Sketched Isotropic Gaussian Regularization (SIGReg) discourages collapse by matching random feature projections to a standard Gaussian~\cite{balestriero2025lejepa}. The auxiliary heads are removed after pretraining; an adapter maps $c_t$ to one policy-width temporal token.

\subsection{Task-conditioned threshold retrieval}
\label{sec:retriever}
Uniform sampling may miss brief evidence, whereas top-$k$ retrieval fixes the frame count. We retain frames above a fixed relevance threshold, adapting memory length to the evidence.

\paragraph{Task-guided supervision}
We construct task-guided supervision in two multimodal stages. First, we identify the visual evidence needed for later decisions; then, we localize events associated with that evidence. For ShellGame, we annotate the initial target--cup association, each exchange's start, trajectory, and end, and the final cup configuration. We visually review these keyframes and obtain RoboMME spatial targets from simulator instance masks. Labels are generated offline and restricted to the observed prefix; annotation prompts are not policy inputs.

\paragraph{Relevance and selection}
Frozen SigLIP uses the policy's So400m/14 visual architecture, producing $N=256$ tokens on a $16{\times}16$ grid from each $224{\times}224$ frame~\cite{zhai2023siglip}. A 256-dimensional projection, patch embeddings, and time encodings precede two causal attention blocks that contextualize each patch location over time. A frozen MiniLM encodes the task goal, followed by a learned query projection~\cite{wang2020minilm}. The normalized query $q$ and keys $K_{in}$ define patch scores and log-sum-exp (LSE) frame logits:
\begin{equation}
s_{in}=\tau q^\top K_{in},\qquad
\ell_i=\log\sum_{n=1}^{N}\exp(s_{in})-\log N,
\end{equation}
Here, $i$ indexes frames, $n$ indexes patches, and $\tau$ is a learned scale. At inference, the retained set is
\begin{equation}
\mathcal{I}_t=\{i\in\{0,\ldots,t\}:\ell_i>0.6\},\qquad m_t=|\mathcal{I}_t|.
\end{equation}
The threshold applies to the normalized LSE logit, not its temporal softmax probability. Retained frames remain in chronological order, with no fixed frame count.

The same score grid receives spatial and temporal supervision. Spatial targets are normalized instance-mask occupancies pooled onto the patch grid. Temporal targets assign uniform mass within two frames of each annotated keyframe and the first three execution frames, clipped to $[0,t]$. Cross-entropy across patches and frames gives $\mathcal{L}_{\mathrm R}=\mathcal{L}_{\mathrm{patch}}+\mathcal{L}_{\mathrm{interval}}$. Patch supervision encourages localized relevance; patch scores do not directly weight the spatial fusion.

\subsection{Linear multi-scale fusion}
\label{sec:patch}
Relevant frames must retain the detail needed for target identification. For each retained frame, we mean-pool the frozen SigLIP $16{\times}16$ features to aligned $8{\times}8$, $4{\times}4$, and $2{\times}2$ grids. We retain sixteen base slots, combining each with its four fine children and coarse parent:
\begin{equation}
\widetilde z_{i,p}=P_4z^{(4)}_{i,p}
+P_8[z^{(8)}_{i,p,1};\ldots;z^{(8)}_{i,p,4}]
+P_2z^{(2)}_{i,\pi(p)}.
\end{equation}
Here, $p$ indexes base slots, $\pi(p)$ identifies the coarse parent, and brackets denote concatenation. Independent linear projections map all scales to a shared feature width. Initializing $P_4$ to the identity and the other projections to zero recovers the base representation. Positional encoding and a feature encoder map the fused features to 1,024-dimensional policy tokens. Each retained frame thus contributes sixteen tokens combining coarse context and fine detail; the spatial memory bank contains $16m_t$ tokens in total.

\subsection{Policy integration and training}
\label{sec:staged}
We use the RoboMME $\pi_{0.5}$ policy stack with a Gemma prefix and flow-matching action expert~\cite{dai2026robomme,black2025pi05,lipman2023flow}. Let $M_{\mathrm{patch}}$ and $M_{\mathrm{temp}}$ denote the spatial memory bank and temporal token. For action-expert hidden state $x$ at layer $l$, the two readouts produce
\begin{equation}
\begin{aligned}
u_l={}&\mathrm{CrossAttn}_{p,l}(x,M_{\mathrm{patch}})\\
&+g_l\mathrm{CrossAttn}_{t,l}(x,M_{\mathrm{temp}}),\quad
g_l=\sigma(\alpha_l).
\end{aligned}
\end{equation}
The signal modulates the normalized action representation before its feed-forward layer, with the temporal gate initialized to 0.01.

Stage 1 pretrains the temporal compressor with $\mathcal{L}_{\mathrm C}$, updating GDN and the final two DINOv3 blocks. Stage 2 freezes GDN and trains the retriever and $\pi_{0.5}$ policy with $\mathcal{L}_{\mathrm{action}}+\mathcal{L}_{\mathrm R}$. SigLIP and MiniLM remain frozen; retrieval supervision trains the relevance scores, while the action objective trains the policy and memory integration. Memory states reset between episodes. Temporal memory has fixed dimension, whereas spatial storage and readout cost depend on $m_t$. Scoring all historical frames additionally incurs a cost that depends on prefix length.

\begin{table*}[!t]
\centering
\papertablecaption{\textbf{Component comparisons.} TSR (\%) across seeds 0, 7, and 42; suite rows report mean $\pm$ standard deviation. Base: published FrameSamp+Modul. Bold marks improvements over Base. $\Delta$: percentage-point difference before rounding. Per-seed scores are in the supplement.}{tab:components}
\small
\setlength{\tabcolsep}{9pt}
\begin{tabular}{@{}llrrr@{}}
\toprule
Component / suite & Task & Base & Addition & $\Delta$ (pp) \\
\midrule
GDN / Counting & BinFill & 39.56 & \textbf{43.33} & \textbf{+3.77} \\
 & StopCube & 42.00 & \textbf{57.33} & \textbf{+15.33} \\
 & PickXtimes & 87.33 & \textbf{94.00} & \textbf{+6.67} \\
 & SwingXtimes & 92.00 & \textbf{94.00} & \textbf{+2.00} \\
 & Suite mean & 65.22 & $\mathbf{72.17\pm4.25}$ & \textbf{+6.94} \\
\midrule
Retrieval / Permanence & ButtonUnmask & 25.11 & \textbf{30.67} & \textbf{+5.56} \\
 & VideoUnmask & 32.67 & 31.33 & -1.34 \\
 & VideoUnmaskSwap & 24.44 & \textbf{30.00} & \textbf{+5.56} \\
 & ButtonUnmaskSwap & 18.22 & \textbf{25.33} & \textbf{+7.11} \\
 & Suite mean & 25.11 & $\mathbf{29.33\pm0.76}$ & \textbf{+4.22} \\
\midrule
Linear multi-scale fusion / Reference & PickHighlight & 22.89 & \textbf{25.33} & \textbf{+2.44} \\
 & VideoRepick & 30.44 & 29.33 & -1.11 \\
 & VideoPlaceButton & 60.00 & \textbf{64.67} & \textbf{+4.67} \\
 & VideoPlaceOrder & 32.00 & \textbf{44.00} & \textbf{+12.00} \\
 & Suite mean & 36.33 & $\mathbf{40.83\pm3.62}$ & \textbf{+4.50} \\
\bottomrule
\end{tabular}
\end{table*}

\begin{figure*}[!t]
\centering
\vbox{\hbox{\includegraphics[width=\textwidth,trim=0 255.015625bp 0 0,clip]{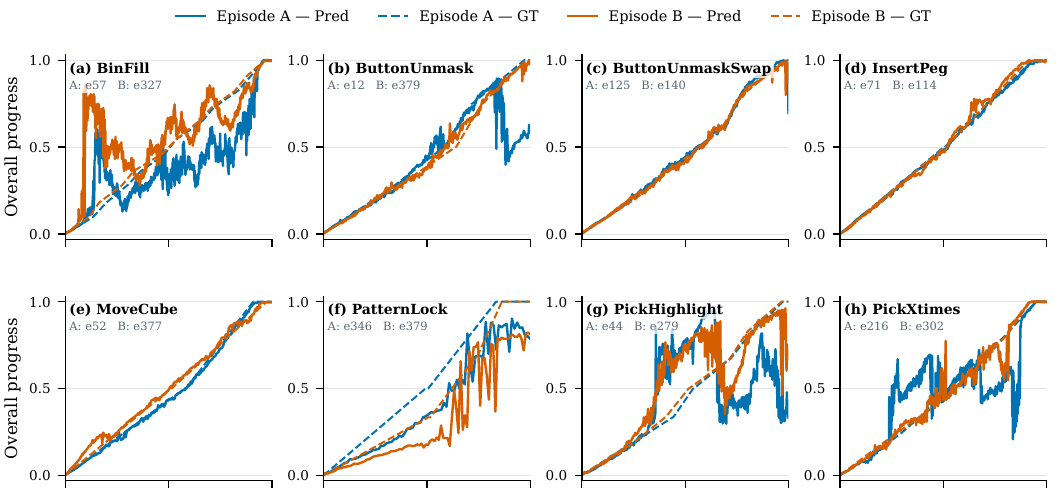}}\nointerlineskip
\hbox{\includegraphics[width=\textwidth,trim=0 0 0 466.68664bp,clip]{figures/overall_progress_pred_vs_gt/overall_progress_pred_vs_gt.pdf}}}
\caption{Current-progress predictions for the first eight tasks in the source ordering, with two episodes per task. Blue and orange identify episodes; solid and dashed curves denote predictions and reference labels. Episode identifiers appear below task names. The horizontal axis is normalized execution time. The readout probes execution-stage information in recurrent memory; the selected traces illustrate task-dependent errors.}
\label{fig:progress-pred-gt}
\end{figure*}

\begin{figure*}[!t]
\centering
\includegraphics[width=\textwidth]{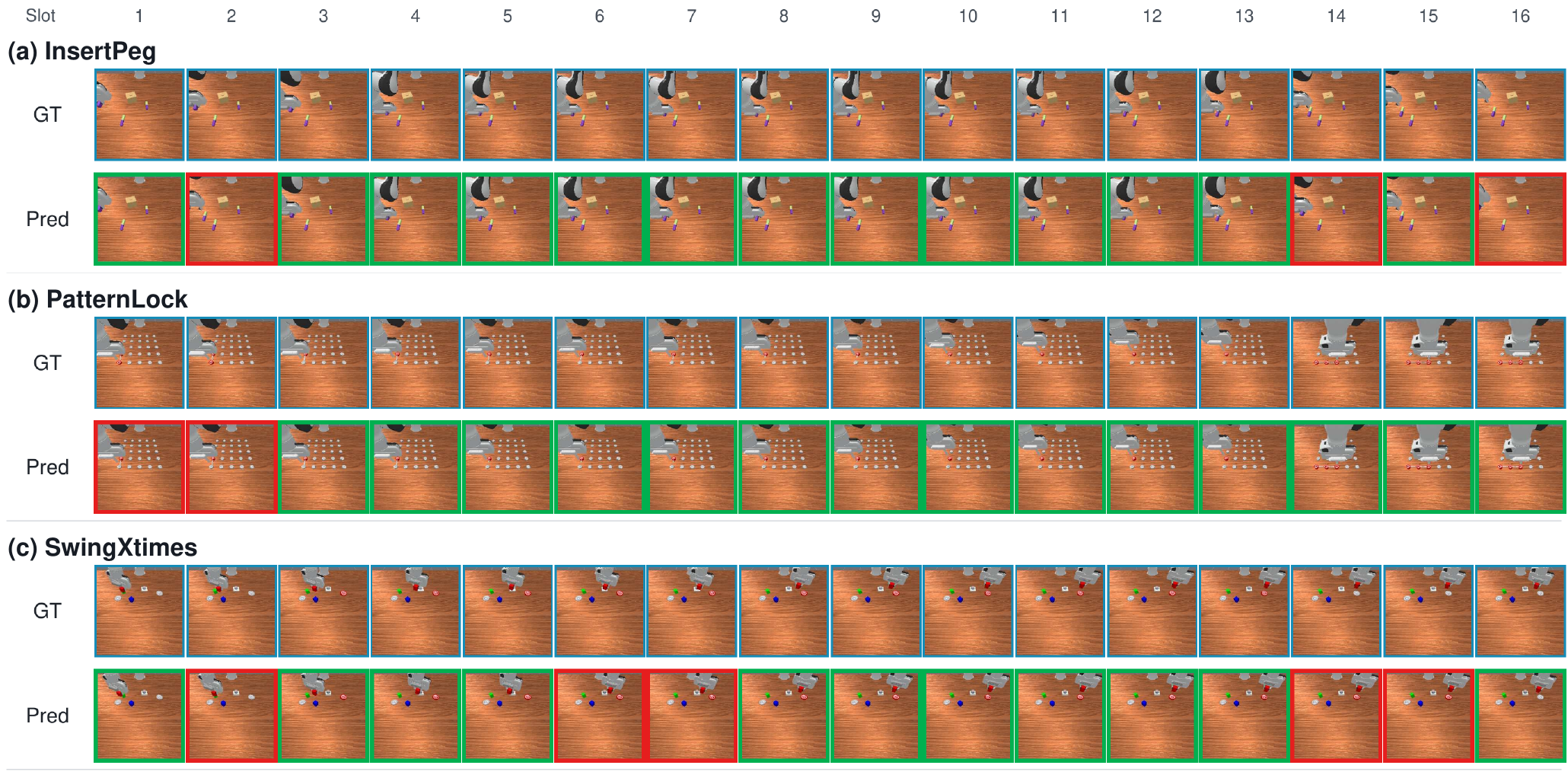}
\caption{Sixteen-slot visual-history comparisons for (a) InsertPeg, (b) PatternLock, and (c) SwingXtimes. Blue and orange borders mark reference (GT) and prediction-associated (Pred) images. On each Pred row, a thick green box marks a scene match and a thick red box marks a mismatch. Columns preserve the supplied slot order. These selected examples illustrate action-conditioned visual-history reconstruction, rather than a dataset-level retrieval metric.}
\label{fig:history-slots}
\end{figure*}

\begin{figure*}[!t]
\centering
\includegraphics[width=\textwidth]{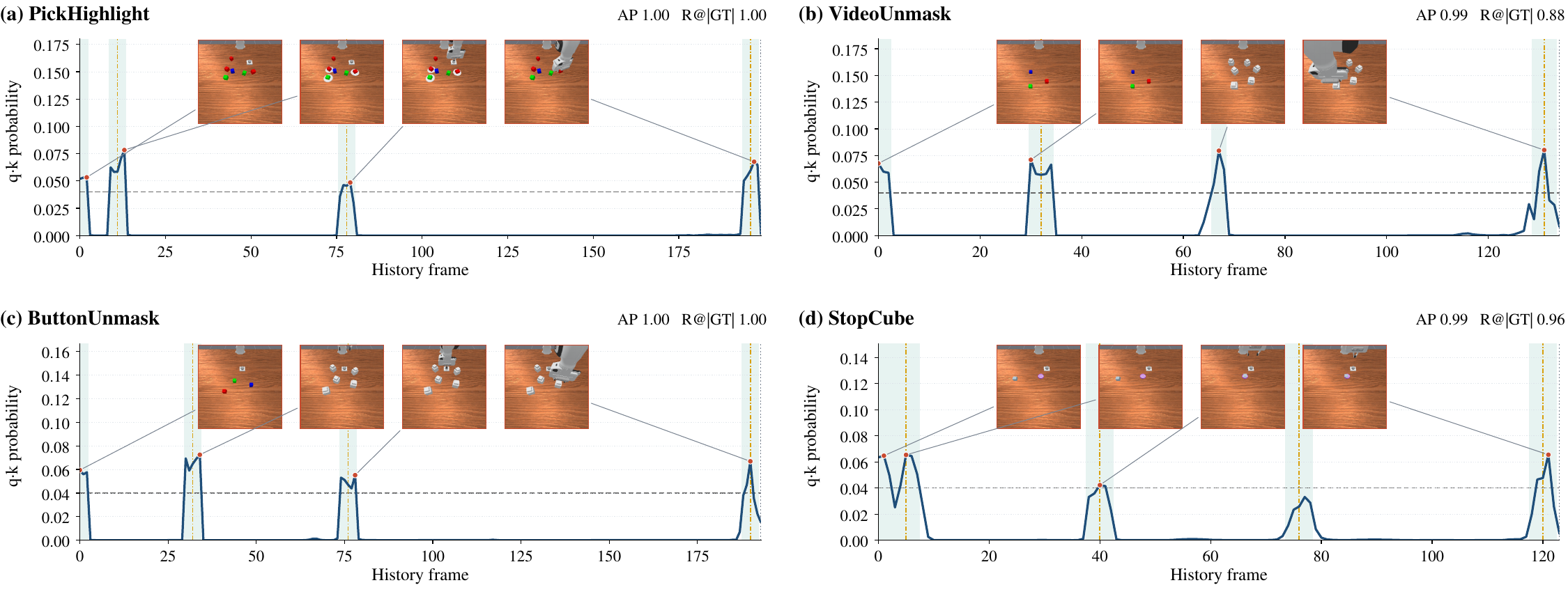}
\caption{Temporal retrieval on four training episodes: (a) PickHighlight, (b) VideoUnmask, (c) ButtonUnmask, and (d) StopCube. Blue curves show normalized scores; shading and gold lines mark positive ranges and keyframe centers. Red markers and thumbnails show peaks above the dashed visualization threshold (0.04). AP denotes average precision; R@$|$GT$|$ denotes recall with as many retrieved frames as positive frames. Inference instead thresholds LSE logits at 0.6 (Section~\ref{sec:retriever}).}
\label{fig:dense-temporal}
\end{figure*}

\begin{table*}[!t]
\centering
\papertablecaption{\textbf{RoboMME evaluation.} TSR (\%) averaged across seeds 0, 7, and 42; suite and overall means weight tasks equally. Bold marks improvements over published FrameSamp+Modul. $\Delta$: percentage-point difference from reported means.}{tab:full}
\renewcommand{\arraystretch}{0.90}
\setlength{\tabcolsep}{6pt}
\small
\begin{tabular}{@{}llrrrrr@{}}
\toprule
Suite & Task & $\pi_{0.5}$ (no memory) & NativeMEM & FrameSamp+Modul & Ours & $\Delta$ (pp) \\
\midrule
Counting & BinFill & 30.00 & 21.33 & 39.56 & \textbf{42.00} & \textbf{+2.44} \\
 & StopCube & 6.67 & 4.00 & 42.00 & \textbf{55.33} & \textbf{+13.33} \\
 & PickXtimes & 42.89 & 24.00 & 87.33 & \textbf{93.33} & \textbf{+6.00} \\
 & SwingXtimes & 35.56 & 40.67 & 92.00 & \textbf{92.67} & \textbf{+0.67} \\
 & Counting mean & 28.78 & 22.50 & 65.22 & \textbf{70.83} & \textbf{+5.61} \\
\midrule
Permanence & ButtonUnmask & 22.22 & 14.00 & 25.11 & \textbf{28.00} & \textbf{+2.89} \\
 & VideoUnmask & 20.44 & 22.67 & 32.67 & \textbf{35.00} & \textbf{+2.33} \\
 & VideoUnmaskSwap & 18.67 & 21.33 & 24.44 & 23.00 & -1.44 \\
 & ButtonUnmaskSwap & 6.67 & 7.33 & 18.22 & \textbf{25.00} & \textbf{+6.78} \\
 & Permanence mean & 17.00 & 16.33 & 25.11 & \textbf{27.75} & \textbf{+2.64} \\
\midrule
Reference & PickHighlight & 11.33 & 23.33 & 22.89 & \textbf{32.67} & \textbf{+9.78} \\
 & VideoRepick & 0.44 & 28.00 & 30.44 & \textbf{35.00} & \textbf{+4.56} \\
 & VideoPlaceButton & 31.11 & 22.00 & 60.00 & \textbf{74.00} & \textbf{+14.00} \\
 & VideoPlaceOrder & 25.78 & 16.00 & 32.00 & \textbf{39.56} & \textbf{+7.56} \\
 & Reference mean & 17.16 & 22.33 & 36.33 & \textbf{45.31} & \textbf{+8.98} \\
\midrule
Imitation & MoveCube & 26.00 & 48.00 & 77.78 & \textbf{83.33} & \textbf{+5.55} \\
 & InsertPeg & 1.56 & 2.00 & 7.56 & \textbf{8.00} & \textbf{+0.44} \\
 & PatternLock & 2.89 & 22.00 & 53.56 & 52.00 & -1.56 \\
 & RouteStick & 4.67 & 12.00 & 66.67 & \textbf{68.67} & \textbf{+2.00} \\
 & Imitation mean & 8.78 & 21.00 & 51.39 & \textbf{53.00} & \textbf{+1.61} \\
\midrule
\multicolumn{2}{l}{Overall mean} & 17.93 & 20.54 & 44.51 & \textbf{49.22} & \textbf{+4.71} \\
\bottomrule
\end{tabular}
\end{table*}

\section{EXPERIMENTS}
\subsection{Setup}
We evaluate component variants on their motivating RoboMME suites, then test the complete model on all sixteen tasks and four real-robot tasks. Stage 2 uses 80,000 steps, batch size 64, and learning rate $5{\times}10^{-5}$ with 10,000 warmup steps. The policy vision encoder is frozen. The retrieval corpus contains 6,400 episodes and 131 task goals.

Task success rate (TSR) uses 50 episodes per task for each evaluation seed, $\{0,7,42\}$. Standard deviations summarize evaluation-seed means; suite means weight tasks equally. FrameSamp+Modul results use the published RoboMME checkpoint aggregation~\cite{dai2026robomme}, which differs from ours. These comparisons measure performance relative to that reference rather than matched architectural effects.

Baselines also include $\pi_{0.5}$ without memory and NativeMEM~\cite{wang2026nativemem}. NativeMEM learns policy-aligned memory tokens, providing a learned-compression baseline alongside the sampled-history baseline FrameSamp+Modul. Our NativeMEM implementation uses 30,000 tokenizer-training steps followed by 60,000 policy-training steps; the policy loss reached approximately $5{\times}10^{-4}$. Supplementary material preserves per-seed component scores, extended training details, and additional spatial-memory comparisons.

\subsection{Component comparisons}
\label{sec:components}
Each variant changes one memory operation within the FrameSamp+Modul interface (Table~\ref{tab:components}). GDN adds a temporal token alongside sampled spatial memory. Retrieval replaces uniform frame selection with threshold retention, using the base spatial features. Linear fusion modifies sampled-frame features, without GDN or threshold retrieval.

\paragraph{Recurrent execution context}
Adding GDN tests stage-dependent decisions: Counting TSR increased by 6.94 percentage points over the published baseline, improving all four tasks. StopCube improved most (+15.33 points), consistent with recurrent context supporting different stopping decisions at visually similar stages.

\paragraph{Task-conditioned selective access}
Replacing uniform sampling tests adaptive access to brief historical evidence. Threshold retrieval increased Permanence TSR by 4.22 percentage points, improving three tasks, most on ButtonUnmaskSwap (+7.11 points). VideoUnmask declined by 1.34 points, while the suite-level gain supports task-conditioned access to historical evidence.

\paragraph{Multi-scale spatial detail}
We test multi-scale features for reference-dependent decisions at fixed token cost. Linear fusion increased Reference TSR by 4.50 percentage points. VideoPlaceOrder improved most (+12.00 points), whereas VideoRepick declined by 1.11 points. The overall gain is consistent with multi-scale features preserving cues for demonstrated-target identification.

\subsection{Analysis of Learned Memory and Retrieval}
\paragraph{Execution-stage readout}
We probe execution-stage information with the auxiliary progress readout. The progress head obtained mean squared error (MSE) 0.03430 over 46,800 valid frames from ten episodes per task across sixteen tasks. Errors are averaged within tasks, then equally across tasks; training uses SmoothL1 loss (Section~\ref{sec:gdn}).

The traces in Fig.~\ref{fig:progress-pred-gt} track the reference closely on InsertPeg and MoveCube. BinFill and PickHighlight show greater fluctuations, and PatternLock underestimates terminal progress. Overall, the readouts indicate that the recurrent state contains information correlated with execution progress.

\paragraph{Action-conditioned visual history}
We probe whether the compressed state supports reconstruction of visual history beyond scalar progress. Figure~\ref{fig:history-slots} compares sixteen reference and prediction-associated images for each of InsertPeg, PatternLock, and SwingXtimes. Scene annotations mark 38 matches and 10 mismatches across 48 displayed slots. These qualitative examples illustrate the historical visual information recoverable from the compressed state under action conditioning.

\paragraph{Temporal and spatial localization}
We evaluate alignment with annotated evidence on the test set. Temporal Hit@1 measures whether the highest-scoring frame lies in a positive window; spatial Hit@1 tests whether the highest-scoring patch intersects the target mask. Average precision (AP) evaluates ranking beyond the first prediction. Temporal and spatial Hit@1 were 96.88\% and 95.00\%, with AP of 96.53\% and 92.24\%, respectively. These ranking metrics support task-conditioned localization of annotated evidence at both temporal and spatial resolutions.

Training-set examples in Fig.~\ref{fig:dense-temporal} illustrate how relevance is distributed over time. The four examples show concentrated responses around annotated evidence, with multiple peaks across each observed history. Their visualization threshold applies to probabilities, whereas inference thresholds normalized LSE logits.

\subsection{Full-benchmark comparison}
\label{sec:full}
We test the combined memory across sixteen RoboMME tasks (Table~\ref{tab:full}). Mean TSR reached 49.22\%, versus 44.51\% for FrameSamp+Modul, 20.54\% for NativeMEM, and 17.93\% without memory. Fourteen tasks and all four suite means improved over FrameSamp+Modul. Reference gained most (+8.98 percentage points), consistent with the value of detailed historical evidence; Counting improved by 5.61 points, consistent with execution-context support.

The benefits were not additive. Reference exceeded the linear-fusion-only variant, but Counting and Permanence remained below their corresponding single-component variants. VideoUnmaskSwap and PatternLock declined by 1.44 and 1.56 percentage points.

NativeMEM improved ten tasks over the no-memory policy but declined on six, including three Counting tasks. Compressed memory alone did not consistently improve performance under this recipe.

\begin{figure}[!t]
\centering
\includegraphics[width=\columnwidth]{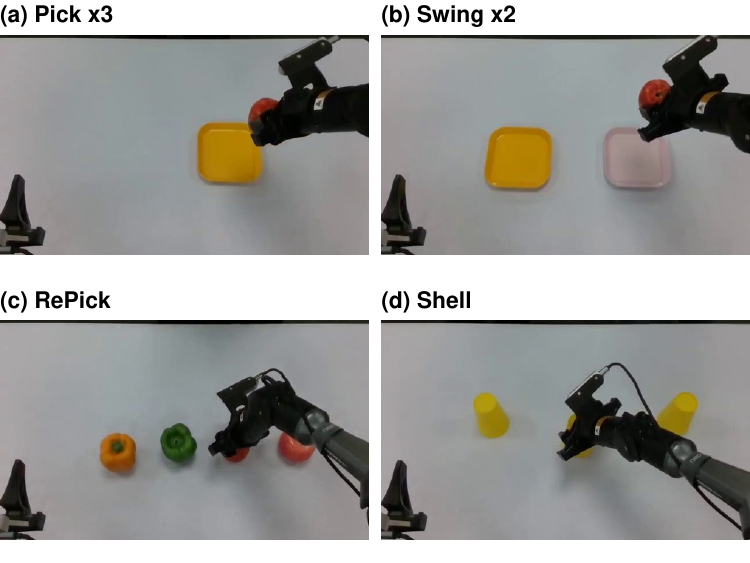}
\caption{Real-robot task examples from recorded demonstrations: (a) repeated picking, (b) repeated swinging, (c) revisiting demonstrated objects, and (d) selecting a cup after swaps. Overhead views show robot execution; these frames illustrate the task settings, not policy-evaluation outcomes.}
\label{fig:real-tasks}
\end{figure}

\subsection{Real-robot evaluation}
\label{sec:real}
We collected 100 demonstrations per task on a Piper dual-arm platform (Fig.~\ref{fig:real-tasks}). Pick$\times3$ and Swing$\times2$ test repeated-action counting; RePick tests demonstrated object order; Shell tests target tracking through three cup swaps. Overhead and left-wrist cameras provide current observations, with overhead history supplying memory.

All methods use the same 400 demonstrations. Each method is evaluated with 30 rollouts per task and matched initial states. Checkpoints are selected at training loss approximately 0.0013; this criterion does not equalize training steps across methods.

\par\noindent\begin{minipage}[t]{\columnwidth}
\papertablecaption{\textbf{Real-robot evaluation.} TSR (\%) over 30 rollouts per task. Bold marks gains over FrameSamp+Modul; Avg. weights tasks equally.}{tab:realrobot}
\centering
\setlength{\tabcolsep}{2.4pt}
{\footnotesize
\begin{tabular}{@{}lccccc@{}}
\toprule
Method & Pick$\times3$ & Swing$\times2$ & RePick & Shell & Avg. \\
\midrule
$\pi_{0.5}$ base & 20.0 & 6.7 & 3.3 & 30.0 & 15.0 \\
NativeMEM & 20.0 & 30.0 & 40.0 & 36.7 & 31.7 \\
FrameSamp+Modul & 26.7 & 36.7 & 33.3 & 53.3 & 37.5 \\
Ours & \textbf{80.0} & \textbf{93.3} & \textbf{36.7} & 40.0 & \textbf{62.5} \\
\bottomrule
\end{tabular}
}
\end{minipage}\par\medskip

The complete model achieved 62.5\% mean TSR, exceeding FrameSamp+Modul by 25.0 percentage points (Table~\ref{tab:realrobot}). Gains concentrated on Pick$\times3$ and Swing$\times2$, with 24/30 and 28/30 successes. Both revisit similar observations at stages requiring different subsequent actions. Together with the Counting comparison, this pattern supports the value of execution context for repeated-action tasks.

RePick reached 36.7\% (11/30), only one success above FrameSamp+Modul, whereas Shell declined from 53.3\% to 40.0\%. Shell requires maintaining target identity through three cup swaps. This task motivates extending frame retrieval with explicit identity tracking across interactions.

\section{DISCUSSION AND CONCLUSION}
We introduced dual-stream memory that combines recurrent execution context with task-conditioned access to historical visual evidence. Component comparisons support the intended memory operations on their motivating suites; the complete policy improves coverage across sixteen benchmark tasks. Real-robot gains concentrate on repeated actions, consistent with execution context helping distinguish visually similar stages.

The results establish performance gains under the reported protocols. Matched full-system ablations would clarify individual stream contributions, while broader task and hardware coverage would assess generalization beyond the current platform.

A concrete next step is to couple bounded retrieval with explicit cross-frame identity propagation. This would address the tracking requirement exposed by cup swaps while controlling storage and readout costs that grow with retained-frame count.

\FloatBarrier


\begin{thebibliography}{00}
\bibitem{brohan2023rt2} B. Zitkovich \emph{et al.}, ``{RT-2}: Vision-language-action models transfer web knowledge to robotic control,'' in \emph{Proc. 7th Conf. Robot Learn. (CoRL)}, ser. PMLR, vol. 229, pp. 2165--2183, 2023.
\bibitem{kim2024openvla} M. J. Kim \emph{et al.}, ``{OpenVLA}: An open-source vision-language-action model,'' in \emph{Proc. 8th Conf. Robot Learn. (CoRL)}, ser. PMLR, vol. 270, pp. 2679--2713, 2025.
\bibitem{ghosh2024octo} D. Ghosh \emph{et al.}, ``Octo: An open-source generalist robot policy,'' in \emph{Proc. Robot.: Sci. Syst. XX (RSS)}, 2024, doi: 10.15607/RSS.2024.XX.090.
\bibitem{black2025pi0} K. Black \emph{et al.}, ``$\pi_0$: A vision-language-action flow model for general robot control,'' in \emph{Proc. Robot.: Sci. Syst. XXI (RSS)}, 2025, doi: 10.15607/RSS.2025.XXI.010.
\bibitem{black2025pi05} K. Black \emph{et al.}, ``$\pi_{0.5}$: A vision-language-action model with open-world generalization,'' in \emph{Proc. 9th Conf. Robot Learn. (CoRL)}, ser. PMLR, vol. 305, pp. 17--40, 2025.
\bibitem{brohan2022rt1} A. Brohan \emph{et al.}, ``{RT-1: Robotics Transformer for Real-World Control at Scale},'' arXiv:2212.06817, 2022.
\bibitem{oxe2023} {Open X-Embodiment Collaboration} \emph{et al.}, ``{Open X-Embodiment: Robotic Learning Datasets and RT-X Models},'' arXiv:2310.08864, 2023.
\bibitem{chi2023diffusion} C. Chi \emph{et al.}, ``{Diffusion Policy: Visuomotor Policy Learning via Action Diffusion},'' arXiv:2303.04137, 2023.
\bibitem{zhao2023act} T. Z. Zhao, V. Kumar, S. Levine, and C. Finn, ``{Learning Fine-Grained Bimanual Manipulation with Low-Cost Hardware},'' arXiv:2304.13705, 2023.
\bibitem{shi2026memoryvla} H. Shi \emph{et al.}, ``{MemoryVLA}: Perceptual-cognitive memory in vision-language-action models for robotic manipulation,'' in \emph{Proc. Int. Conf. Learn. Represent. (ICLR)}, 2026, arXiv:2508.19236.
\bibitem{torne2026mem} M. Torne \emph{et al.}, ``{MEM}: Multi-scale embodied memory for vision language action models,'' arXiv:2603.03596, 2026.
\bibitem{wang2026nativemem} Z. Wang, M. Shi, C. Ni, J. Yang, M. Li, Z. Su, T. Lin, and H. Li, ``{NativeMEM}: Native memory compression for long-horizon robotic manipulation,'' arXiv:2607.06678, 2026.
\bibitem{mees2022calvin} O. Mees, L. Hermann, E. Rosete-Beas, and W. Burgard, ``{CALVIN: A Benchmark for Language-Conditioned Policy Learning for Long-Horizon Robot Manipulation Tasks},'' arXiv:2112.03227, 2021.
\bibitem{liu2023libero} B. Liu \emph{et al.}, ``{LIBERO: Benchmarking Knowledge Transfer for Lifelong Robot Learning},'' arXiv:2306.03310, 2023.
\bibitem{dai2026robomme} Y. Dai, H. Fu, J. Lee, Y. Liu, H. Zhang, J. Yang, C. Finn, N. Fazeli, and J. Chai, ``{RoboMME}: Benchmarking and understanding memory for robotic generalist policies,'' in \emph{Proc. Int. Conf. Mach. Learn. (ICML)}, 2026, arXiv:2603.04639.
\bibitem{jaegle2021perceiver} A. Jaegle \emph{et al.}, ``{Perceiver: General Perception with Iterative Attention},'' arXiv:2103.03206, 2021.
\bibitem{alayrac2022flamingo} J.-B. Alayrac \emph{et al.}, ``{Flamingo: a Visual Language Model for Few-Shot Learning},'' arXiv:2204.14198, 2022.
\bibitem{vaswani2017attention} A. Vaswani \emph{et al.}, ``{Attention Is All You Need},'' arXiv:1706.03762, 2017.
\bibitem{dai2019transformerxl} Z. Dai \emph{et al.}, ``{Transformer-XL: Attentive Language Models Beyond a Fixed-Length Context},'' arXiv:1901.02860, 2019.
\bibitem{rae2019compressive} J. W. Rae, A. Potapenko, S. M. Jayakumar, and T. P. Lillicrap, ``{Compressive Transformers for Long-Range Sequence Modelling},'' arXiv:1911.05507, 2019.
\bibitem{bulatov2022rmt} A. Bulatov, Y. Kuratov, and M. S. Burtsev, ``Recurrent memory transformer,'' \emph{Adv. Neural Inf. Process. Syst.}, vol. 35, pp. 11079--11091, 2022.
\bibitem{katharopoulos2020linear} A. Katharopoulos, A. Vyas, N. Pappas, and F. Fleuret, ``Transformers are {RNNs}: Fast autoregressive transformers with linear attention,'' in \emph{Proc. 37th Int. Conf. Mach. Learn. (ICML)}, ser. PMLR, vol. 119, pp. 5156--5165, 2020.
\bibitem{gu2023mamba} A. Gu and T. Dao, ``Mamba: Linear-time sequence modeling with selective state spaces,'' in \emph{Proc. 1st Conf. Lang. Model. (COLM)}, 2024.
\bibitem{dao2024mamba2} T. Dao and A. Gu, ``Transformers are {SSMs}: Generalized models and efficient algorithms through structured state space duality,'' in \emph{Proc. 41st Int. Conf. Mach. Learn. (ICML)}, ser. PMLR, vol. 235, pp. 10041--10071, 2024.
\bibitem{yang2024deltanet} S. Yang, B. Wang, Y. Zhang, Y. Shen, and Y. Kim, ``Parallelizing linear transformers with the delta rule over sequence length,'' \emph{Adv. Neural Inf. Process. Syst.}, vol. 37, pp. 115491--115522, 2024, doi: 10.52202/079017-3668.
\bibitem{yang2025gateddelta} S. Yang, J. Kautz, and A. Hatamizadeh, ``Gated delta networks: Improving {Mamba2} with delta rule,'' in \emph{Proc. Int. Conf. Learn. Represent. (ICLR)}, 2025, arXiv:2412.06464.
\bibitem{sun2024ttt} Y. Sun \emph{et al.}, ``Learning to (learn at test time): {RNNs} with expressive hidden states,'' in \emph{Proc. 42nd Int. Conf. Mach. Learn. (ICML)}, ser. PMLR, vol. 267, pp. 57503--57522, 2025.
\bibitem{jiang2026robottt} Y. Jiang \emph{et al.}, ``{RoboTTT}: Context scaling for robot policies,'' arXiv:2607.15275, 2026.
\bibitem{dosovitskiy2020vit} A. Dosovitskiy \emph{et al.}, ``{An Image is Worth 16x16 Words: Transformers for Image Recognition at Scale},'' arXiv:2010.11929, 2020.
\bibitem{caron2021dino} M. Caron \emph{et al.}, ``{Emerging Properties in Self-Supervised Vision Transformers},'' arXiv:2104.14294, 2021.
\bibitem{oquab2023dinov2} M. Oquab \emph{et al.}, ``{DINOv2}: Learning robust visual features without supervision,'' \emph{Trans. Mach. Learn. Res.}, 2024.
\bibitem{simeoni2025dinov3} O. Sim{\'e}oni \emph{et al.}, ``{DINOv3},'' \emph{Trans. Mach. Learn. Res.}, 2026.
\bibitem{radford2021clip} A. Radford \emph{et al.}, ``Learning transferable visual models from natural language supervision,'' in \emph{Proc. 38th Int. Conf. Mach. Learn. (ICML)}, ser. PMLR, vol. 139, pp. 8748--8763, 2021.
\bibitem{zhai2023siglip} X. Zhai, B. Mustafa, A. Kolesnikov, and L. Beyer, ``Sigmoid loss for language image pre-training,'' in \emph{Proc. IEEE/CVF Int. Conf. Comput. Vis. (ICCV)}, pp. 11941--11952, 2023, doi: 10.1109/ICCV51070.2023.01100.
\bibitem{lin2016fpn} T.-Y. Lin \emph{et al.}, ``{Feature Pyramid Networks for Object Detection},'' arXiv:1612.03144, 2016.
\bibitem{liu2021swin} Z. Liu \emph{et al.}, ``{Swin Transformer: Hierarchical Vision Transformer using Shifted Windows},'' arXiv:2103.14030, 2021.
\bibitem{ronen2023mixedres} T. Ronen, O. Levy, and A. Golbert, ``Vision transformers with mixed-resolution tokenization,'' in \emph{Proc. IEEE/CVF Conf. Comput. Vis. Pattern Recognit. Workshops (CVPRW)}, pp. 4613--4622, 2023, doi: 10.1109/CVPRW59228.2023.00486.
\bibitem{shridhar2022peract} M. Shridhar, L. Manuelli, and D. Fox, ``{Perceiver-Actor: A Multi-Task Transformer for Robotic Manipulation},'' arXiv:2209.05451, 2022.
\bibitem{balestriero2025lejepa} R. Balestriero and Y. LeCun, ``{LeJEPA}: Provable and scalable self-supervised learning without the heuristics,'' arXiv:2511.08544, 2025.
\bibitem{wang2020minilm} W. Wang, F. Wei, L. Dong, H. Bao, N. Yang, and M. Zhou, ``{MiniLM}: Deep self-attention distillation for task-agnostic compression of pre-trained transformers,'' \emph{Adv. Neural Inf. Process. Syst.}, vol. 33, pp. 5776--5788, 2020.
\bibitem{lipman2023flow} Y. Lipman, R. T. Q. Chen, H. Ben-Hamu, M. Nickel, and M. Le, ``Flow matching for generative modeling,'' in \emph{Proc. Int. Conf. Learn. Represent. (ICLR)}, 2023, arXiv:2210.02747.
\end{thebibliography}
\end{document}